\def\year{2021}\relax
\documentclass[letterpaper]{article} 
\usepackage{aaai21}  
\usepackage{times}  
\usepackage{helvet} 
\usepackage{courier}  
\usepackage[hyphens]{url}  
\usepackage{graphicx} 
\usepackage{natbib}  
\usepackage{caption} 
\usepackage{amsmath, amssymb, amsthm}
\usepackage[linesnumbered,ruled,vlined]{algorithm2e}
\usepackage{quoting}
\quotingsetup{vskip=0pt}
\usepackage{nicefrac}       
\usepackage{bbm}
\usepackage{enumitem}
\usepackage{booktabs}       
\usepackage{subcaption}
\usepackage{appendix}
\newtheorem{property}{Property}

\usepackage{colortbl}
\usepackage{refcount}
\usepackage{comment}
\usepackage{dsfont}

\newcommand{\update}{\textit{MemoryUpdate}}
\newcommand{\retrieval}{\textit{MemoryRetrieval}}
\newcommand\Tstrut{\rule{0pt}{2.6ex}}         
\newcommand\Bstrut{\rule[-0.9ex]{0pt}{0pt}}   

\newcommand{\clear}[1]{}

\title{Online Class-Incremental Continual Learning with Adversarial Shapley Value}
\author {

    Dongsub Shim\textsuperscript{\rm 1}\thanks{Authors contributed equally.},
    Zheda Mai\textsuperscript{\rm 1}\footnotemark[1]\thanks{Corresponding author}, 
    Jihwan Jeong\textsuperscript{\rm 1}\footnotemark[1], Scott Sanner\textsuperscript{\rm 1}\\
    Hyunwoo Kim\textsuperscript{\rm 2}, Jongseong Jang\textsuperscript{\rm 2}\\
}
\affiliations {
    \textsuperscript{\rm 1} University of Toronto, \\
    \textsuperscript{\rm 2} LG AI Research\\
    (dongsub.shim, zheda.mai)@mail.utoronto.ca, 
    (jhjeong, ssanner)@mie.utoronto.ca, 
    (hwkim, j.jang)@lgresearch.ai
    
}

\begin{document}

\maketitle

\begin{abstract}
%
As image-based deep learning becomes pervasive on every device from cell phones to smart watches, there is a growing need to develop methods that continually learn from data while minimizing memory footprint and power consumption.  
While memory replay techniques have shown exceptional promise for this task of \emph{continual learning}, the best method for selecting which buffered images to replay is still an open question.  
In this paper, we specifically focus on the online class-incremental setting where a model needs to learn new classes continually from an online data stream.  To this end, we 
contribute a novel
Adversarial Shapley value scoring method 
that scores memory data samples according to their ability to preserve latent decision boundaries for previously observed classes (to maintain learning stability and avoid forgetting) while interfering with latent decision boundaries of current classes being learned (to encourage plasticity and optimal learning of new class boundaries).  Overall, we observe that our proposed ASER method provides competitive or improved performance compared to state-of-the-art replay-based continual learning methods on a variety of datasets.
\end{abstract}
\section{Introduction}
Image-based deep learning is a pervasive but computationally expensive and memory intensive task.  Yet the need for such deep learning on personal devices to preserve privacy, minimize communication bandwidth, and maintain real-time performance necessitates the development of methods that can continuously learn from streaming data while minimizing memory storage and computation footprint.  
However, a well-documented defect of deep neural networks that prevents it from learning continually is called \emph{catastrophic forgetting}~\cite{CF} --- the inability of a network to perform well in previously seen tasks after learning new tasks. 
To address this challenge, the field of \emph{continual learning} (CL) studies the problem of learning from a (non-iid) stream of data, with the goal of preserving and extending the acquired knowledge over time.


Many existing CL approaches use a \emph{task incremental} setting where data arrives one task (i.e., set of classes to be identified) at a time and the model can utilize task identity during both training and testing~\cite{Kirkpatrick2017, Li2016, gem}. Specifically, a common practice in this setting is to assign a separate output layer (head)
for each task; then the model just needs to classify labels within a task, which is known as multi-head evaluation~\cite{riemannian}. However, this setting requires additional supervisory signals at test time --- namely the task identity --- to select the corresponding head, which obviates its use when the task label is unavailable. 
In this work,  we consider a more realistic but difficult setting, known as \emph{online class-incremental}, where a model needs to learn new classes continually from an online data stream (each sample is seen only once). In contrast to the task incremental setting, this setting adopts the single-head evaluation, where the model needs to classify all labels without task identity. Moreover, we focus on image classification, a common application where this setting is used~\citep{mir, gss, dirichlet, onlineImbalance}.

\begin{figure*}[t!]
    \centering
    \includegraphics[width=0.9\linewidth]{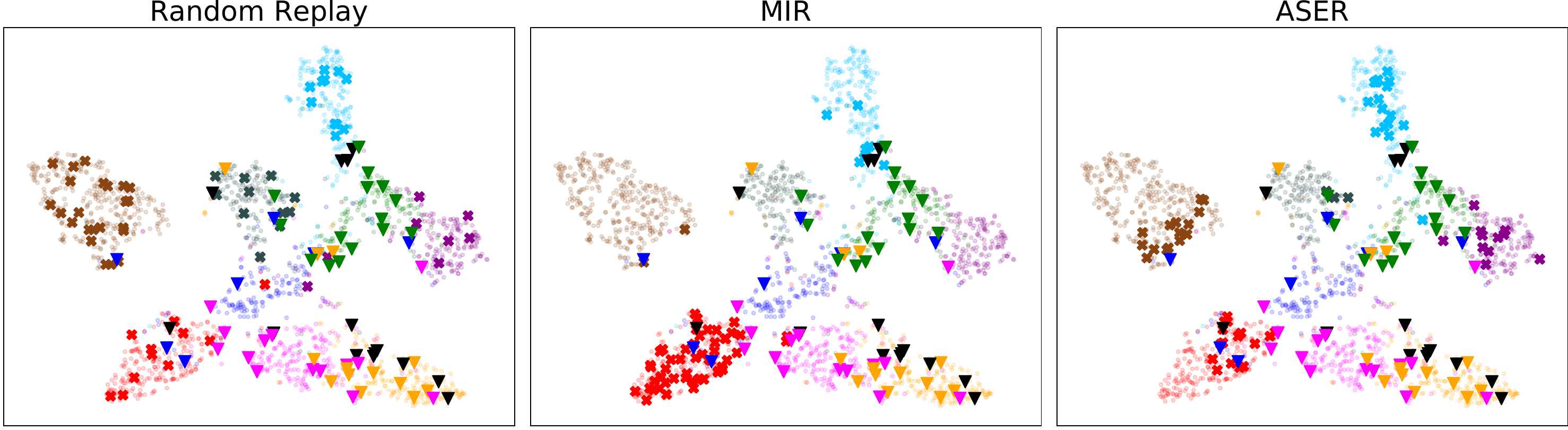}
    \caption{2D t-SNE \cite{tsne} visualization of CIFAR-100 data embeddings and their class labels (different colors) showing
    current task samples (triangle), memory samples (pale circle) and retrieved memory samples for rehearsal (bold x). For each point, we obtain the latent embedding from reduced ResNet18~\cite{agem}.
Note that Random Replay~\cite{tiny} distributes its retrieved samples non-strategically, MIR disproportionately selects seemingly redundant samples in a single -- apparently most interfered -- class (red), 
    whereas ASER strategically retrieves memory samples that are representative of different classes in memory but also adversarially located near class boundaries and current task samples.}
    \label{fig:tsne}
\end{figure*}

Current CL methods can be taxonomized into three major categories: regularization-based, parameter isolation, and memory-based methods \cite{Parisi2019, survey}. Regularization-based methods incorporate an additional penalty term into the loss function to penalize the update of critical model parameters~\cite{Kirkpatrick2017, SI, mas, laplace}. 
Other regularization-based methods imposed knowledge distillation techniques to 
penalize the feature drift on previous tasks~\cite{Li2016, largescale, encoder}.
Parameter isolation methods assign per-task parameters to bypass interference by expanding the network and masking parameters to prevent forgetting~\cite{packnet, dirichlet, expandable}.  Memory-based methods deploy a memory buffer to store a subset of data from previous tasks. The samples from the buffer can be either used to constrain the parameter updates such that the loss on previous tasks cannot increase \cite{agem, gem}, or 
simply for replay to prevent forgetting~\cite{rebuffi2017icarl, tiny}.

Regularization methods only protect the model's ability to classify within a task and thus they do not work well in our setting where the ability to discriminate among classes from different tasks is crucial~\cite{lesort2019regularization}. Also, most parameter isolation methods require task identity during inference, which violates our setting. Therefore in this work, we consider the replay approach which has shown to be successful and efficient for the online class-incremental setting~\cite{gss, mir}. 
Since the memory buffer is the only place to store data from previous tasks, a key question for replay-based methods is \emph{how to update and retrieve memory samples when new data arrives?} For example, \citet{tiny} proposed a simple but strong baseline that randomly updates and retrieves samples, while the highly effective Maximally
Interfered Retrieval (MIR) method~\cite{mir} chooses replay samples whose loss most increases after a current task update.  However, if we visualize 
the latent space of retrieved memory samples chosen by each method in Figure~\ref{fig:tsne}, we see that the methods mentioned above fail to strategically select samples that both preserve existing memory-based class boundaries while protecting against current task samples that interfere with these boundaries (detailed discussion in caption of Figure~\ref{fig:tsne}).



 

We address the deficiencies observed above by proposing a novel replay-based method called Adversarial Shapley value Experience Replay (ASER). ASER is inspired by the Shapley value (SV)~\cite{shapley1953value} used in cooperative game theory to fairly distribute total gains to all players --- in our CL setting, we use the SV to determine the contribution of memory samples to learning performance~\cite{datashapley,knnsv,datavaluation}.  We also introduce an adversarial perspective of SV for CL memory retrieval that aims to score memory samples according to their preservation of decision boundaries for ``friendly'' samples in the memory buffer (to maintain learning stability and avoid forgetting) and their interference with ``opponent'' samples from the current task that disrupt existing memory-based class boundaries (to encourage plasticity and optimal learning). 
Through extensive experiments on three commonly used benchmarks in the CL literature, we demonstrate that ASER  provides competitive or improved performance compared to  state-of-the-art replay-based methods, especially when the memory buffer size is small.
\section{Continual Learning}

\subsection{Problem Definition}

\paragraph{Online Class-Incremental Learning}
Following the recent CL literature \cite{mir,gss, dirichlet, onlineImbalance}, we consider the online supervised class-incremental learning setting where a model needs to learn new classes continually from an online data stream (each sample is seen only once). More concretely, a neural network classifier $f:\mathbb{R}^d\mapsto\mathbb{R}^C$, parameterized by $\theta$ will receive input batches $B_n^t$ of size $b$ from task$_t$. Task$_t$ consists of classes that the classifier has never seen before in task$_{1:t}$.
Moreover, we adopt the single-head evaluation setup \cite{riemannian} where the classifier has no access to task identity during inference and hence must choose among all labels. Our goal is to train the classifier $f$ to continually learn new classes from the data stream without forgetting.

\paragraph{Metrics}
Since the goal of CL is to continually acquire new knowledge while preserving existing learning, we use two standard metrics in the CL literature to measure performance: \emph{average accuracy} for overall performance and \emph{forgetting} to measure how much acquired knowledge the algorithm has forgotten~\cite{riemannian,tiny}. In \textbf{Average Accuracy}, $a_{i,j}$ is the accuracy evaluated on the held-out test set of task $j$ after training the network from task 1 to $i$. In \textbf{Average Forgetting}, 
$f_{i, j}$ represents how much the model forgets about task $j$ after being trained on task $i$. For $T$ tasks:
\begin{align*}
\text{Average Accuracy} (A_{T})&=\frac{1}{T} \sum_{j=1}^{T} a_{T, j}\\
\text{Average Forgetting} (F_{T})&=\frac{1}{T-1} \sum_{j=1}^{T-1} f_{T, j} \\
\text{where}~~f_{i, j}&=\max _{l \in\{1, \cdots, i-1\}} a_{l, j}-a_{i, j}
\end{align*}

\subsection{Experience Replay Methods}
\paragraph{Experience Replay (ER)}
The research of ER and the important role of replay buffers has been well-established in the reinforcement learning area \cite{rolnick2019experience, foerster2017stabilising}. Recently, ER has been widely applied in supervised CL learning tasks \cite{mer, mir, gss}. Compared with the simplest baseline model that fine-tunes the parameters based on the new task without any measures to prevent forgetting, ER makes two simple modifications: (1) it stores a subset of the samples from past tasks in a memory buffer $\mathcal{M}$ of limited size $M$; (2) it concatenates the incoming minibatch $B_n$ with another minibatch $B_\mathcal{M}$ of samples selected from the memory buffer. Then, it simply takes a SGD step with the combined batch, followed by an online update of the memory. A generic ER algorithm is presented in Algorithm \ref{alg:algo_er}.

What differentiates various replay-based methods are the \retrieval{} in line 3 and the \update{} in line 5. Although there exists another stream of replay methods that utilize a generative model to produce virtual samples instead of using a memory buffer \cite{generative}, recent research has demonstrated the limitations of such approaches with convolutional neural networks in datasets such as CIFAR-10 \cite{mir, gen_cl_timothee}. Hence, we focus on the memory-based approach in our work.

Basic \textbf{ER} is a simple but strong baseline that applies reservoir sampling in \update{} and random sampling in \retrieval{}. Despite its simplicity, recent research has shown that naive ER outperforms many specifically designed CL approaches with and without a memory buffer \cite{tiny}. 

\begin{algorithm}
\caption{Generic ER-based method}
\label{alg:algo_er}
\SetCustomAlgoRuledWidth{0.49\textwidth}
\SetAlgorithmName{Algorithm}{}{}
    \SetKwInOut{Input}{Input~}
    \SetKwInOut{Output}{Output}
    \SetKwInput{Initialize}{Initialize}
\Input{Batch size $b$, Learning rate $\alpha$}
\Initialize{Memory $\mathcal{M}\leftarrow \{\}*M$; Parameters $\theta$; Counter $n\leftarrow 0$ }
    \For{$t\in\{1,\dots,T\}$}{
        \For{$B_n\sim D_t$}{
        $B_\mathcal{M}\!\!\leftarrow\!$ \retrieval{}($B_n,\! \mathcal{M}$)\\
        $\theta\leftarrow~\text{SGD}(B_n\cup B_\mathcal{M},\theta, \alpha)$ \\
        $\mathcal{M}\leftarrow$ \update{}$(B_n, \mathcal{M})$\\
        $n\leftarrow n + b$
        }
    }
     \textbf{return} $\theta$
\end{algorithm}

\paragraph{Maximally-interfered Retrieval (MIR)}
MIR is a recently proposed method aiming to improve the \retrieval{} strategy \cite{mir}. MIR chooses replay samples according to loss increases given the estimated parameters update based on the newly arrived data. However, samples with significant loss increases tend to be similar in the latent space, which may lead to redundancy in the retrieved data, as shown in Figure \ref{fig:tsne}.  Like ER, MIR uses reservoir sampling for the \update{}.

\paragraph{Gradient-based Sample Selection (GSS)}
Different from MIR, GSS pays attention to the \update{} strategy \cite{gss}. Specifically, it tries to diversify the gradients of the samples in the memory buffer. Like ER,  GSS uses random sampling in \retrieval{}.

\section{Efficient Computation of Shapley Value via KNN Classifier}

When we return to Figure~\ref{fig:tsne} and analyze the latent embeddings of memory samples, we observe the natural clustering effect of classes in the embedding space, which has been well-observed previously in the deep learning literature~\cite{nn-embedding, cnn-embedding}.  On account of this, we observe that some samples may indeed be more important than others in terms of preserving what the neural network has learned.  For example, data from one class that are near the boundary with data from another class in some sense act as sentinels to guard the decision boundary between classes.  This suggests the following question: \emph{how can we value data in the embedded space in terms of their contribution to accurate classification?}

Given that the embedding plot of Figure~\ref{fig:tsne} suggests that a new data point is likely to take the classification of its nearest neighbors in the embedding space, we could rephrase this question as asking how much each data point in memory contributes to correct classification from the perspective of a K-Nearest Neighbors (KNN) classifier.  Fortunately, 
the existing research literature already provides both a precise and efficient answer to this question viewed through the lens of Shapley data valuation for KNN classifiers~\cite{knnsv, datashapley, datavaluation}.  Before we cover this solution, we first pause to recap the purpose of Shapley values.

\paragraph{Shapley Value (SV) for Machine Learning}
The SV~\cite{shapley1953value,shapley} was originally proposed in cooperative game theory to decide the share of total gains for each player in a coalition. The SV has a set of mathematical properties that make it appealing to many applications: \emph{group rationality}, \emph{fairness}, and \emph{additivity}. Conversely, it can be shown that the SV is the \emph{only} allocation scheme that satisfies these three properties. 

In the context of machine learning, the SV has been used to estimate the individual contribution of data points to the performance of a trained model \emph{in the context of all other data}~\cite{datashapley,datavaluation}. 
Formally, let $N$ denote the number of data points and $I=\{1,\dots,N\}$ be the associated index set. Then, each datum is interpreted as a player of a cooperative game with the goal of maximizing test-time performance. Let $v(S)$ define a utility function of the ML model over a subset $S\subset I$ on which the model is trained. Then, the SV of a data point of index $i$ with the utility $v(S)$ is the following:
\begin{equation}
 s(i) = \frac{1}{N}\sum_{S\subseteq{I\backslash{\{i\}}}} \frac{1}{\binom{N-1}{|S|}}{[v(S \cup \{i\}) - v(S)]} \label{eq:def_sv}
\end{equation}
Intuitively, when we consider every possible subset of data points, $s(i)$ measures the average marginal improvement of utility given by the sample $i$. By setting the utility as test accuracy in ML classification tasks, the SV can discover how much of the test accuracy is attributed to a training instance. 

\paragraph{Efficient KNN Shapley Value Computation}
Specific to our requirements for data valuation in this paper, recent work has developed an efficient method for SV computation in a KNN classification framework~\cite{knnsv}.  This is a critical innovation since the direct powerset-based computation of the SV requires
$O(2^N)$ evaluations for general, bounded utility functions. 
Furthermore, each evaluation involves training an ML model with a given subset of data ($S$). This is prohibitive in most modern deep learning applications, not to mention online CL with neural networks. 
As shown in \cite{knnsv} and summarized below, the exact KNN-SV can be computed in $O(N\log N)$.

Let $(\mathbf{x}_j^\text{ev}, y_j^\text{ev})$ denote an \emph{evaluation point} and $D_c=\{(\mathbf{x}_i,y_i)\}_{i=1}^{N_c}$ a \emph{candidate set}, where $y_j^\text{ev}$ and $y_i$ are labels. We compute the KNN-SVs of all examples in $D_c$ w.r.t. the evaluation point with the utility function \eqref{eq:singleutility}. The KNN utility function over a subset $S\subset D_c$ measures the likelihood of correct classifications:
\begin{align}
    v_{j,\text{KNN}}(S) = \frac
    {1}{K}\sum_{k=1}^{\min(K, |S|)}\mathds{1}[y_{\alpha_{k}(S)} = \ y_j^{\text{ev}}] \label{eq:singleutility}
\end{align}
where $\alpha_k(S)$ is the index of the $k$th closest sample (from $\mathbf{x}_j^\text{ev}$) in $S$ based on some distance metric. Each sample $i$ is assigned a KNN-SV \textemdash $~s_j(i)~$\textemdash \space that represents the average marginal contribution of the instance to the utility. Due to the additivity of SV, we obtain the KNN-SV of a candidate sample w.r.t. the evaluation set ($D_e=\{(\mathbf{x}^\text{ev}_j,y^\text{ev}_j)\}_{j=1}^{N_e}$) by taking the average: $s_\text{avg}(i) = \nicefrac{1}{N_e}\sum_{j=1}^{N_e}s_j(i)$.

\eqref{eq:farthestsv} and \eqref{eq:recur_knnsv} show how to recursively compute the exact KNN-SVs of samples in $D_c$ w.r.t. $(\mathbf{x}_{j}^{ \text{ev}}, y_{j}^{\text{ev}})\in D_e$ starting from $\mathbf{x}_{\alpha_{N_c}}$ (the farthest point from $\mathbf{x}_{j}^{ \text{ev}}$) \cite{knnsv}:
\begin{align}
    s_j(\alpha_{N_c}) &= \frac{\mathds{1}[y_{\alpha_{N_c}}=  y_j^{\text{ev}}]}{N_c} \label{eq:farthestsv} \\
    s_j(\alpha_m) &= s_j(\alpha_{m+1}) + \nonumber \\ 
    & \frac{\mathds{1}[y_{\alpha_m}= y_j^{\text{ev}}] - \mathds{1}[y_{\alpha_{m+1}}=\  y_j^{\text{ev}}]}{K}\frac{{\min}(K, m)}{m} \label{eq:recur_knnsv}
\end{align}
Here, $s_j(\alpha_m)$ is the KNN-SV of the $m$th closest candidate sample from $\mathbf{x}_j^\text{ev}$. Note that the dependency on the utility $v$ is suppressed as $v_\text{KNN}$ is always used. We refer readers to \cite{knnsv} for detailed derivation of these results.

\section{Adversarial Shapley Value Experience Replay (ASER)} \label{sec:aser}


We have now affirmatively answered how to value data in the embedded space in terms of its contribution to accurate classification by leveraging the efficient KNN-SV computation.  Equipped with this powerful global data valuation algorithm, we now present our novel ER method dubbed Adversarial Shapley value ER (ASER) that leverages the SV for both \retrieval{} and \update.  


A key insight with our ASER approach for \retrieval{} is that we need to balance the competing needs at the crux of CL, i.e., we need to retrieve memory samples 
for replay that prevent forgetting while also finding samples that maximally interfere with the incoming batch $B_n$ to ensure 
plasticity in learning.  This leads us not only to leverage a cooperative notion of the SV (where higher SV is better) as it relates to $\mathcal{M}$ but also an adversarial notion of the SV as it relates to $B_n$ (where lower -- and, in fact, negative -- SVs indicate interference).
In addition ASER also adopts a cooperative SV approach to the \update{} process.

Formally, we can view a neural network classifier ($f$) as two separate parts: a feature extractor ($f_\text{ext}: \mathbb{R}^d\mapsto\mathbb{R}^h$) and a fully connected neural classifier ($f_\text{cl}:\mathbb{R}^h\mapsto\mathbb{R}^C$), where $h$ is the dimensionality of the latent space $\mathcal{X}^l$. We implicitly define a KNN classifier and use the Euclidean distance in $\mathcal{X}^l$. Then, by \eqref{eq:farthestsv}-\eqref{eq:recur_knnsv}, we can compute the KNN-SVs of candidate samples w.r.t. evaluation samples. 


As previously noted, ER's performance depends on deciding what to store in memory (i.e., \update{}) and what to replay from memory (i.e., \retrieval{}). 
One key desiderata is that we want samples in $\mathcal{M}$ as well as $B_n$ to be well-separated by $f_\text{ext}$ in the latent space. To this end, we target two types of samples in $\mathcal{M}$ for retrieval: those near the samples in $B_n$ but have different labels (Type 1); those that are representative of samples in the memory (Type 2). Training with samples in Type 1 encourages the model to learn to differentiate current classes from previously seen classes. Samples in Type 2 help retain latent decision boundaries for previously observed classes.

We ground our intuition as to how samples interfere and cluster with each other in the latent space based on two properties of the KNN-SV. Given a candidate sample $i\in D_c$ and an evaluation set $D_e$, the KNN-SV of the point $i$ w.r.t. an evaluation point $j\in D_e$, i.e. $s_j(i)$, satisfies the following (see Appendix\footnote{\label{fn:appendix} Please find the appendix in our extended version on arXiv. Link: https://arxiv.org/abs/2009.00093} A for proof):
\begin{itemize}[leftmargin=*, wide=0pt]
{
  \item \textbf{Property 1.}  $s_j(i) > 0$ if and only if $y_i = y^\text{ev}_j$. Also, $s_j(i)=0$ only when $S=\{i'|y_{i'}=y^\text{ev}_j, ~\forall i'\in \{i+1,\dots,N_c\}\}=\emptyset$.
  \item \textbf{Property 2.} $|s_j(m)|$ is a non-increasing function of $m$ for $m$ such that $y_m=y_j^\text{ev}$. Similarly, $|s_j(n)|$ is a non-increasing function of $n$ for $n$ such that $y_n\ne y_j^\text{ev}$. And for $m\ge K$, $|s_j(m)|-|s_j(m')|>0$ holds for $m<m'$, where $m'$ is the smallest index with $\mathds{1}(y_m=y_j^\text{ev})=\mathds{1}(y_{m'}=y_j^\text{ev})$, if there exists $l\in (m, m')$ such that $\mathds{1}(y_l=y_j^\text{ev})\ne\mathds{1}(y_{m}=y_j^\text{ev})$. In other words, as $i$ gets closer to the evaluation point $j$, $|s_j(i)|$ cannot decrease for points with the same $\mathds{1}(y_i=y_j^\text{ev})$, and for $i\ge K$, it can only increase when there exist more than one differently labeled points.
  
  }
\end{itemize}

The first property states that a candidate sample $i$ has a \emph{positive KNN-SV} if it has the same label as the evaluation point being considered (\emph{cooperative}); the sample will have a \emph{negative KNN-SV} if its label is different than the evaluation point (\emph{adversarial}). By combining both properties, we note: 
\vspace{1mm}
\begin{quoting}[leftmargin=4ex]
    If $s_j(i)$ is large, the candidate $i$ is close to the evaluation point $j$ in the latent space ($\mathcal{X}^l$) and has the same label ($y_i=y_j^\text{ev}$). On the other hand, if $s_j(i)$ is a negative value of large magnitude, then $i$ is close to $j$, yet has a different label ($y_i\neq y_j^\text{ev}$).  Thus, we conjecture that a good data candidate $i$ has \emph{high positive SV} for memory $\mathcal{M}$ and \emph{negative SV with large magnitude} for the current input task $B_n$.
\end{quoting}
\vspace{5mm}

When we consider the whole evaluation set, we take the mean $s_{D_e}(i)=\nicefrac{1}{|D_e|}\cdot\sum_{j\in D_e}s_j(i)$, and the above analysis still holds in average. 
%
Therefore, by examining the KNN-SVs of candidate samples, we can get a sense of how they are distributed with respect to the evaluation set in $\mathcal{X}^l$.
Then, we define the \textbf{adversarial SV (ASV)} that encodes the Type 1 \& 2 criteria 
\begin{align}
    \textbf{ASV}(i) = \max_{j\in S_\text{sub}} s_j (i) - \min_{k \in B_n} s_k(i),
    \label{eq:adv_obj_minmax}
\end{align}
as well as a ``softer'' mean variation \textbf{ASV$_\mu$}
\begin{align}
    \textbf{ASV}_\mu(i) = \frac{1}{|S_\text{sub}|}{\sum_{j\in S_\text{sub}} s_j(i)} - \frac{1}{b}{\sum_{k\in B_n} s_{k}(i)},
    \label{eq:adv_obj_mean}
\end{align}
where $i \in \mathcal{M} \setminus S_\text{sub}$ and $S_\text{sub}$ is constructed by subsampling some number of examples from $\mathcal{M}$ such that it is balanced in terms of the number of examples from each class. This prevents us from omitting any latent decision boundaries of classes in the memory. Note that $S_\text{sub}$ is used as the evaluation set in the first term, whereas the input batch $B_n$ forms the evaluation set in the latter term. The candidate set is $\bar{\mathcal{M}}=\mathcal{M}\setminus S_\text{sub}$, and we retrieve samples of size $b_\mathcal{M}$ from the set that have the highest \textbf{ASV}s (Algorithm \ref{alg:algo_aser}). We denote our ER method using the score \textbf{ASV} \eqref{eq:adv_obj_minmax} as ASER, while ASER$_\mu$ uses \textbf{ASV}$_\mu$ \eqref{eq:adv_obj_mean} instead. For computational efficiency, we randomly subsample $N_c$
candidates from $\bar{\mathcal{M}}$.

\begin{algorithm}
\caption{ASER \textit{MemoryRetrieval}}
\label{alg:algo_aser}
\SetCustomAlgoRuledWidth{0.43\textwidth}
\DontPrintSemicolon
\SetAlgorithmName{Algorithm}{}{}
    \SetKwInOut{Input}{Input~}
    \SetKwInOut{Output}{Output}
    \SetKwInput{Initialize}{Initialize}
\small
\Input{Memory batch size $b_\mathcal{M}$\\
Input batch $B_n$; Candidate size $N_c$;\\ Subsample size $N_\text{sub}$;\\ 
Feature extractor $f_\text{ext}$}
    $S_\text{sub}\overset{N_\text{sub}}{\sim} \mathcal{M}$ 
    \tcp*{get evaluation set}
    $D_c~\overset{N_c}{\sim}~ \mathcal{M}\setminus S_\text{sub}$
    \tcp*{get candidate set}
    \footnotesize
    \tcc{get latent embeddings}
    \small
    $L_{B_n}, L_{S_\text{sub}}, L_{D_c}\leftarrow f_\text{ext}(B_n), f_\text{ext}(S_\text{sub}), f_\text{ext}(D_c)$\\
    \For{$i \in D_c$}{
    \For{$j\in S_\text{sub}$}{
    $s_j(i)\leftarrow$ KNN-SV($L_{D_c},L_{S_\text{sub}} $) as per \eqref{eq:farthestsv}, \eqref{eq:recur_knnsv}
    
    }
    \For{$k\in B_n$}{
    $s_k(i)\leftarrow $ KNN-SV($L_{D_c},L_{B_n} $) as per \eqref{eq:farthestsv}, \eqref{eq:recur_knnsv}
    }
    $score(i)\leftarrow$ \textbf{ASV}$(i)$ as per \eqref{eq:adv_obj_minmax} or \eqref{eq:adv_obj_mean}
    }
    $B_\mathcal{M}\leftarrow$ $b_\mathcal{M}$ samples with largest $score(\cdot)$
    
\textbf{return} $B_\mathcal{M}$
\end{algorithm}

Note that both ASER methods do not greedily retrieve samples with the smallest distances to either $S_\text{sub}$ or $B_n$. This is because for a single evaluation point $j$, $s_j(\alpha_m)=s_j(\alpha_{m+1})$ when $y_{\alpha_m}=y_{\alpha_{m+1}}$. So, a few points can have the same score even if some of them are farther from the evaluation point. This is in contrast to a pure distance-based score where the closest point gets the highest score.
In Appendix\footnotemark[\getrefnumber{fn:appendix}] B, we show that our method outperforms pure distance-based methods, proving the effectiveness of the global way in which the SV scores candidate data based on the KNN perspective.

We summarize our method in Algorithm \ref{alg:algo_aser}, and compare it with other state-of-the-art ER methods on multiple challenging CL benchmarks in Section \ref{sec:experiment}.

\paragraph{Memory Update Based on KNN-SV}
For \update{}, we find that samples with high KNN-SV promote clustering effect in the latent space. Therefore, they are useful to store in the memory, which aligns with the original meaning of the SV. More concretely, we subsample $S_\text{sub}\sim \mathcal{M}$ and compute $\nicefrac{1}{|S_\text{sub}|}\sum_{j\in S_\text{sub}} s_j(i)$ for $i\in \bar{\mathcal{M}}\cup B_n$. Then, we replace samples in $\bar{\mathcal{M}}$ having smaller average KNN-SVs than samples in $B_n$ with the input batch samples. 

We use KNN-SV \update{} for ASER throughout experiments in Section \ref{sec:experiment}, while the ablation analysis of different variations with random \update{} or random \retrieval{} (both random retrieval and update reduces to ER) is presented in Appendix\footnotemark[\getrefnumber{fn:appendix}] C. Note that ASER with KNN-SV \update{} performs competitively or better than the variations, underscoring the importance of SV-based methods for both \update{} and \retrieval{}.


%
%

\section{Experiments} \label{sec:experiment}
To test the efficacy of ASER and its variant ASER$_\mu$, we evaluate their performance by comparing them with several state-of-the-art CL baselines. We begin by reviewing the benchmark datasets, baselines we compared against and our experiment setting. We then report and analyze the result to validate our approach.

\subsection{Datasets}
\noindent\textbf{Split CIFAR-10} splits the CIFAR-10 dataset \cite{cifar} into 5 different tasks with non-overlapping classes and 2 classes in each task, similarly as in \cite{mir}. 

\noindent\textbf{Split CIFAR-100} is constructed by splitting the CIFAR-100 dataset \cite{cifar} into 10 disjoint tasks, and each task has 10 classes.

\noindent\textbf{Split miniImagenet} consists of splitting the miniImageNet dataset \cite{miniimagenet} into 10 disjoint tasks, where each task contains 10 classes


The detail of datasets, including the general information of each dataset,  class composition and the number of samples in training, validation and test sets of each task is presented in Appendix\footnotemark[\getrefnumber{fn:appendix}] D.

{
\begin{table*}[ht!]
    \small
    \centering
    \begin{tabular}{ c@{\kern0.7em} @{}c@{\kern0.7em} @{}c@{\kern0.7em} @{}c@{\kern0.7em} | @{\kern0.7em}c@{\kern0.7em} @{}c@{\kern0.7em} @{}c@{\kern0.7em} | @{\kern0.7em}c@{\kern0.7em} @{}c@{\kern0.7em} @{}c@{\kern0.7em} @{}c} 
    \toprule
    Method &M=1k& M=2k &  M=5k\Bstrut & M=1k& M=2k &  M=5k &  M=0.2k &  M=0.5k & M=1k \\
    \hline\hline
    \Tstrut
    iid online & $14.7\pm0.6$&$14.7\pm0.6$&$14.7\pm0.6$&$20.5\pm0.4$&$20.5\pm0.4$&$20.5\pm0.4$&$62.9\pm1.5$&$62.9\pm1.5$&$62.9\pm1.5$ \\
    iid offline & $42.4\pm0.4$&$42.4\pm0.4$&$42.4\pm0.4$&$47.4\pm0.3$&$47.4\pm0.3$&$47.4\pm0.3$&$79.7\pm0.4$&$79.7\pm0.4$&$79.7\pm0.4$ \\
    \midrule
    AGEM & $7.0\pm0.4$ &$7.1\pm0.5$ &$6.9\pm0.7$&$9.5\pm0.4$&$9.3\pm0.4$&$9.7\pm0.3$&$22.7\pm1.8$&$22.7\pm1.9$&$22.6\pm0.7$\\
    ER & $8.7\pm0.4$&$11.8\pm0.9$&$16.5\pm0.9$&$11.2\pm0.4$&$14.6\pm0.4$&$20.1\pm0.8$&$26.4\pm1.0$&$32.2\pm1.4$&$38.4\pm1.7$\\
    EWC & $3.1\pm0.3$ &$3.1\pm0.3$ &$3.1\pm0.3$&$4.8\pm0.2$&$4.8\pm0.2$&$4.8\pm0.2$& $17.9\pm0.3$&$17.9\pm0.3$&$17.9\pm0.3$\\
    fine-tune&$4.3\pm0.2$&$4.3\pm0.2$&$4.3\pm0.2$&$5.9\pm0.2$&$5.9\pm0.2$&$5.9\pm0.2$&$17.9\pm0.4$&$17.9\pm0.4$&$17.9\pm0.4$ \\
    GSS&$7.5\pm0.5$&$10.7\pm0.8$&$12.5\pm0.4$&$9.3\pm0.2$&$10.9\pm0.3$&$15.9\pm0.4$&$26.9\pm1.2$&$30.7\pm1.2$&$40.1\pm1.4$\\
    MIR&$8.1\pm0.3$&$11.2\pm0.7$&$15.9\pm1.6$&$11.2\pm0.3$&$14.1\pm0.2$&$21.2\pm0.6$&$28.3\pm1.6$&$35.6\pm1.2$&$42.4\pm1.5$\\
    \midrule
    ASER&$\mathbf{11.7\pm0.7}$&$\mathbf{14.4\pm0.4}$&$\mathbf{18.2\pm0.7}$&$12.3\pm0.4$&$14.7\pm0.7$&$20.0\pm0.6$&$27.8\pm1.0$&$36.2\pm1.1$&$43.1\pm1.2$\\
    ASER$_\mu$&$\mathbf{12.2\pm0.8}$&$\mathbf{14.8\pm1.1}$&$\mathbf{18.2\pm1.1}$&$\mathbf{14.0\pm0.4}$&$\mathbf{17.2\pm0.5}$&$21.7\pm0.5$&$26.4\pm1.5$&$36.3\pm1.2$&$43.5\pm1.4$\\
    \hline
    \bottomrule
    \Tstrut
     &  \multicolumn{3}{c}{(a) Mini-ImageNet\label{accu-mini-imagenet} } & \multicolumn{3}{c}{\label{accu-CIFAR-100} (b) CIFAR-100} & \multicolumn{3}{c}{\label{accu-cifar-10} (c) CIFAR-10}\\
    \end{tabular}
    \label{tab:acc_miniimagenet}

\caption{Average Accuracy (higher is better), M is the memory buffer size. All numbers are the average of 15 runs. ASER$_\mu$ has better performance when M is small and dataset is more complex.\protect\footnotemark}
\label{tab:acc_main}
\end{table*}
}

\subsection{Baselines}
We compare our proposed ASER against several state-of-the-art continual learning algorithms:
\begin{itemize}
      \item\textbf{AGEM} \cite{agem}: Averaged Gradient Episodic Memory, a memory-based method that utilizes the samples in the memory buffer to constrain the parameter updates.  
  \item \textbf{ASER \& ASER$_\mu$}: Our proposed methods. ASER scores samples in the memory with \textbf{ASV} in \eqref{eq:adv_obj_minmax}. ASER$_\mu$ uses the mean variation \textbf{ASV}$_\mu$ in \eqref{eq:adv_obj_mean}.
  \item \textbf{ER} \cite{tiny}: Experience replay, a recent and successful rehearsal method with random sampling in \retrieval{} and reservoir sampling in \update{}.
    \item \textbf{EWC} \cite{Kirkpatrick2017}: Elastic Weight Consolidation, a prior-focused method that limits the update of parameters that were important to the past tasks, as measured by the Fisher information matrix. 
  \item \textbf{GSS} \cite{gss}: Gradient-Based Sample Selection, a \update{}  method that diversifies the gradients of the samples in the replay memory.
  \item \textbf{MIR} \cite{mir}: Maximally Interfered Retrieval, a \retrieval{} method that retrieves memory samples that suffer from an increase in loss given the estimated parameters update based on the current task.
  \item \textbf{iid-online \& iid-offline}: iid-online trains the model with a single-pass through the same set of data, but each mini-batch is sampled iid from the training set. iid-offline trains the model over multiple epochs on the dataset with iid sampled mini-batch. We use 5 epochs for iid-offline in all the experiments as in \cite{mir, gss}. 
  \item \textbf{fine-tune}: As an important baseline in previous work \cite{mir, gss, dirichlet}, it simply trains the model in the order the data is presented without any specific method for forgetting avoidance.
\end{itemize}

\subsection{Experiment Setting}

\footnotetext{The discrepancy of CIFAR-10 result for MIR between the original paper and this work is discussed in Appendix\footnotemark[\getrefnumber{fn:appendix}]~F}

{
\begin{table*}[ht!]
    \small
    \centering
    \begin{tabular}{ c@{\kern0.7em} @{}c@{\kern0.7em} @{}c@{\kern0.7em} @{}c@{\kern0.7em} | @{\kern0.7em}c@{\kern0.7em} @{}c@{\kern0.7em} @{}c@{\kern0.7em} | @{\kern0.7em}c@{\kern0.7em} @{}c@{\kern0.7em} @{}c@{\kern0.7em} @{}c} 
    \toprule
    Method &M=1k& M=2k &  M=5k\Bstrut & M=1k& M=2k &  M=5k &  M=0.2k &  M=0.5k & M=1k \\
    \hline\hline
    \Tstrut
    AGEM & $29.3\pm0.9$ &$30.0\pm0.9$ &$29.9\pm0.8$&$40.4\pm0.7$&$39.7\pm0.8$&$39.8\pm1.0$& $36.1\pm3.8$&$43.2\pm4.2$&$48.1\pm3.0$\\
    ER & $29.7\pm1.3$&$29.2\pm0.9$&$26.6\pm1.1$&$45.0\pm0.5$&$40.5\pm0.8$&$34.5\pm0.8$&$72.8\pm1.7$&$63.1\pm2.4$&$55.8\pm2.6$\\
     EWC & $28.1\pm0.8$ &$28.1\pm0.8$ &$28.1\pm0.8$&$39.1\pm1.2$&$39.1\pm1.2$&$39.1\pm1.2$&$81.5\pm1.4$&$81.5\pm1.4$&$81.5\pm1.4$\\
    fine-tune &$35.6\pm0.9$&$35.6\pm0.9$&$35.6\pm0.9$&$50.4\pm1.0$&$50.4\pm1.0$&$50.4\pm1.0$&$81.7\pm0.7$&$81.7\pm0.7$&$81.7\pm0.7$ \\
    GSS &$29.6\pm1.2$& $27.4\pm1.1$&$29.9\pm1.2$& $46.9\pm0.7$&$42.3\pm0.8$&$39.2\pm0.9$& $75.5\pm1.5$&$65.9\pm1.6$&$54.9\pm2.0$\\
    MIR &$29.7\pm1.0$& $27.2\pm1.1$&$26.2\pm1.4$&$45.5\pm0.8$&$40.4\pm0.6$&$31.4\pm0.6$&$67.0\pm2.6$&$68.9\pm1.7$&$47.7\pm2.9$\\
    \midrule
    ASER &$30.1\pm1.3$& $\mathbf{24.7\pm1.0}$&$\mathbf{20.9\pm1.2}$&$50.1\pm0.6$&$45.9\pm0.9$&$36.7\pm0.8$&$71.1\pm1.8$&$\mathbf{59.1\pm1.5}$&$50.4\pm1.5$\\
    ASER$_\mu$ &$28.0\pm1.3$&$\mathbf{22.2\pm1.6}$&$\mathbf{17.2\pm1.4}$&$45.0\pm0.7$&$\mathbf{38.6\pm0.6}$&$30.3\pm0.5$&$72.4\pm1.9$&$\mathbf{58.8\pm1.4}$&$47.9\pm1.6$\\
    \hline
    \bottomrule
    \Tstrut
     &  \multicolumn{3}{c}{(a) Mini-ImageNet\label{forget-mini-imagenet} } & \multicolumn{3}{c}{\label{forget-CIFAR-100} (b) CIFAR-100} & \multicolumn{3}{c}{\label{forget-cifar-10} (c) CIFAR-10}\\
    \end{tabular}
    \label{tab:forget_miniimagenet}

\caption{Average Forgetting (lower is better). Memory buffer size is M. All numbers are the average of 15 runs.}
\label{tab:forget_main}
\end{table*}
}

\begin{figure}
    \begin{subfigure}{3.3in}
        \centering
        \includegraphics[
        height=3.6cm
        ]{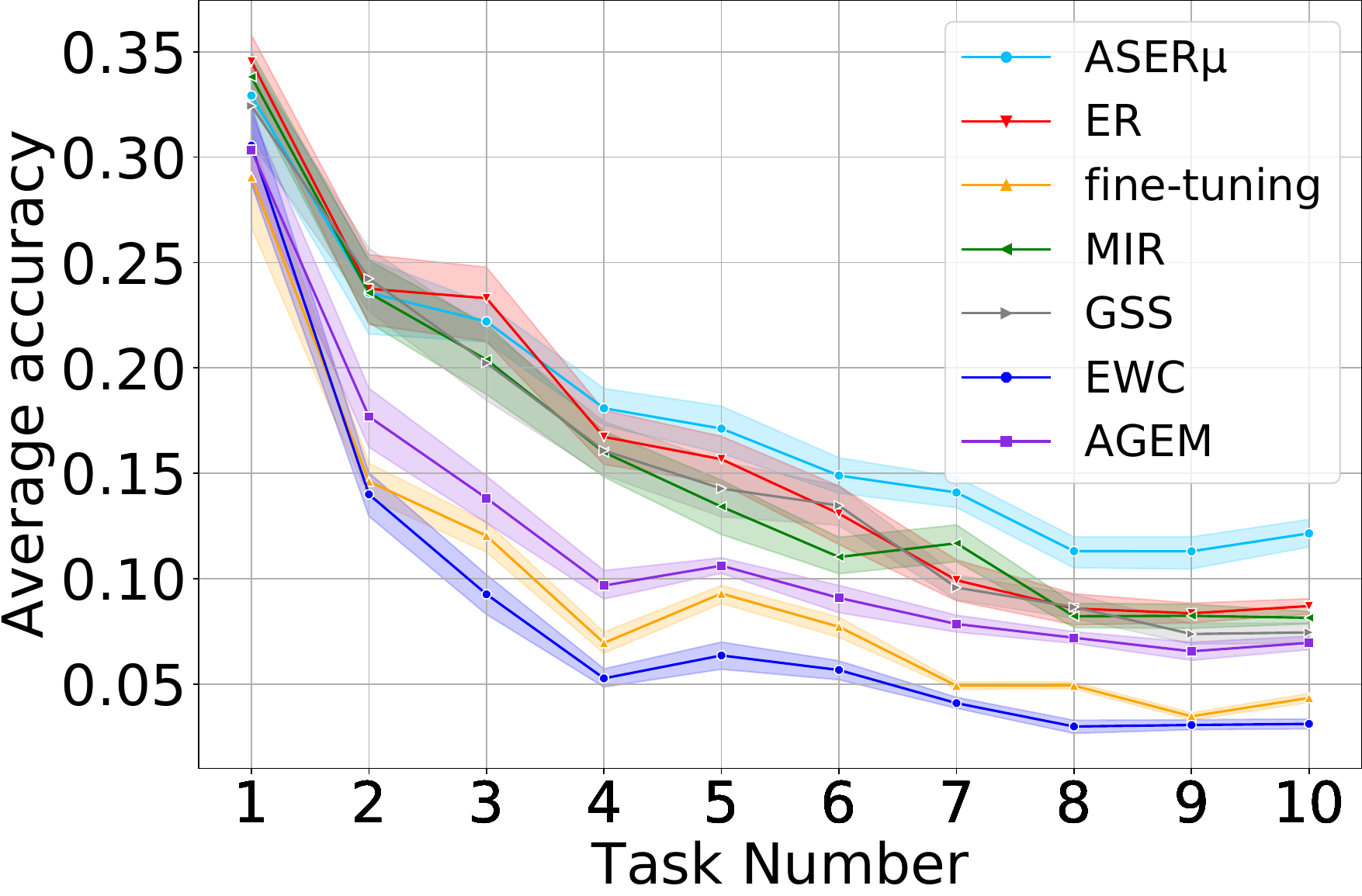}
        \caption{Mini-ImageNet}
        \label{fig:mini}
    \end{subfigure}
    \begin{subfigure}{3.3in}
    \centering
    \includegraphics[
        height=3.6cm
        ]{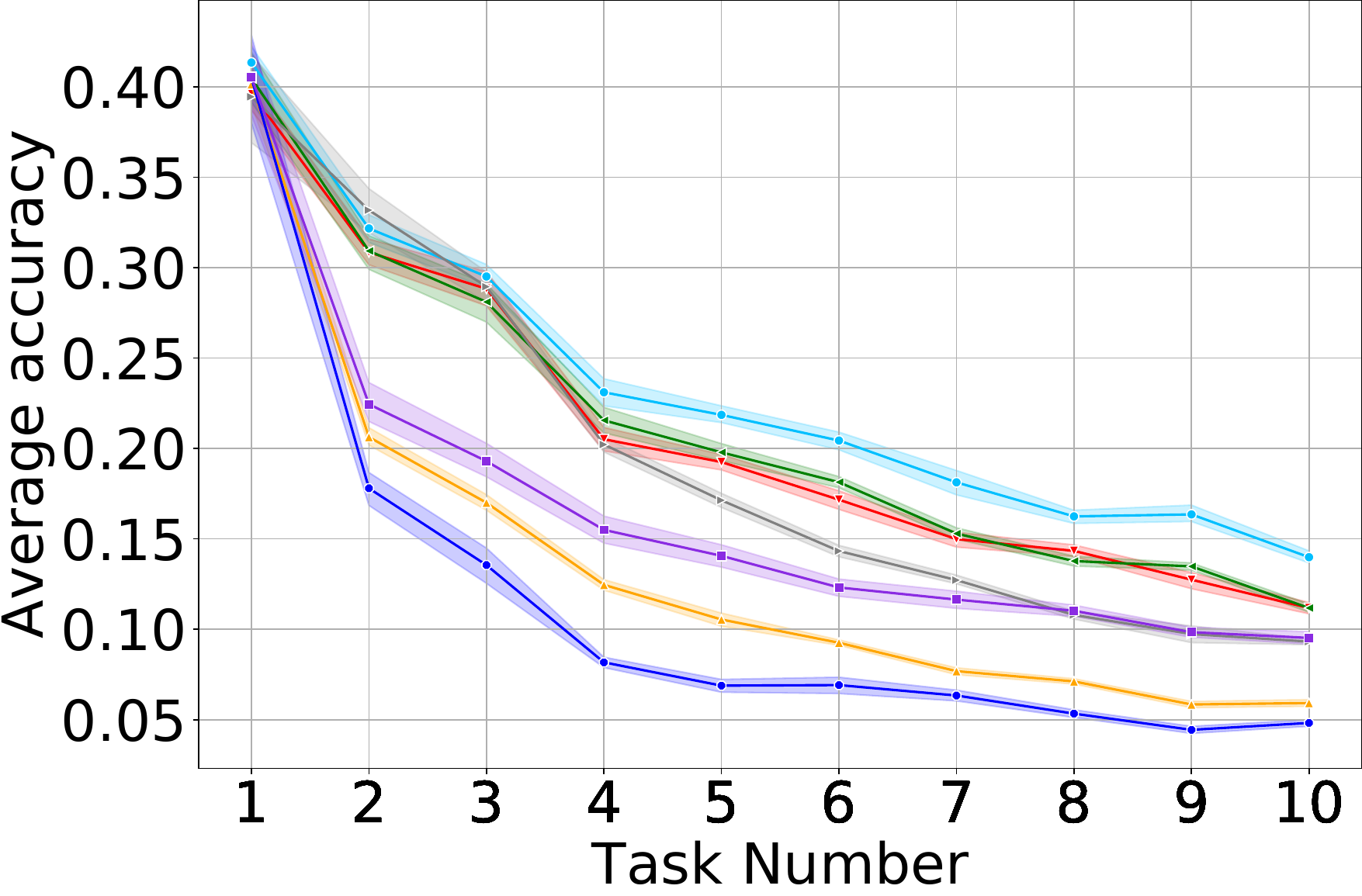}
    \caption{CIFAR-100}
    \label{fig:cifar100}
    \end{subfigure}
    \hfill
    \caption{Average accuracy on observed tasks when M=1k. The shaded region represents the 95\% confidence interval. ASER$_\mu$ outperforms other baselines especially when the model sees more classes (each task contains new classes).}\label{fig:acc}
\end{figure}


\paragraph{Single-head Evaluation}Most of the previous work in CL applied multi-head evaluation \cite{riemannian} where a distinct output head is assigned for each task and the model utilizes the task identity to choose the corresponding output head during test time. But in many realistic scenarios, task identity is not available during test time, so the model should be able to classify labels from different tasks. As in \cite{mir,gss}, we adopt the single-head evaluation setup where the model has one output head for all tasks and is required to classify all labels. Note that the setting we use -- online and single-head evaluation -- is more challenging than many other reported CL settings.

\paragraph{Model}We use a reduced ResNet18, similar to \cite{tiny, gem}, as the base model for all datasets, and the network is trained via cross-entropy loss with SGD optimizer and mini-batch size of 10. The size of the mini-batch retrieved from memory is also set to 10 irrespective of the size of the memory. More details of the experiment can be found in Appendix\footnotemark[\getrefnumber{fn:appendix}]E.

\subsection{Comparative Performance Evaluation}
Table \ref{tab:acc_main} and Table \ref{tab:forget_main} show the average accuracy and average forgetting by the end of the data stream for Mini-ImageNet, CIFAR-100 and CIFAR-10. 
Based on the performance of iid-online and iid-offline, we verify that Mini-ImageNet and CIFAR-100 are more complex than CIFAR-10, even though three datasets have the same number of samples. Overall, ASER and ASER$_\mu$ show competitive or improved performance in three standard CL datasets. Especially, we observe that ASER$_\mu$ outperforms all the state-of-the-art baselines by significant margins in a more difficult setting where memory size is small and dataset is complex. Since the difficulty of the three datasets is different, comparing the absolute accuracy improvement may not be fair. Therefore, percentage improvement\footnote{Percentage improvement is the ratio between absolute improvement and baseline performance. For example, in Mini-ImageNet(M=1k), ASER$_\mu$ improves MIR by $\frac{12.2-8.7}{8.7}=40.2\%$} is more appropriate here. Taking Mini-ImageNet as an example, ASER$_\mu$ improves the strongest baseline by 40.2\% (M=1k), 25.4\% (M=2k) and 10.3\% (M=5k) in terms of percentage improvement. Moreover, as we can see in Figure~\ref{fig:acc}, ASER$_\mu$ is consistently better than other baselines in both datasets. We also note that ASER$_\mu$ generally performs better than ASER. This is because if we use the ASV criterion as in \eqref{eq:adv_obj_minmax}, it has a higher chance that the value is affected by an outlier point in the \emph{evaluation} set. So the \textbf{ASV}$_\mu$ in \eqref{eq:adv_obj_mean} gives a more stable and accurate value in complicated datasets.


Another interesting observation is that ER has very competitive performances. Especially in more complex datasets,  it surpasses GSS and performs similarly as MIR, which proves it to be a simple but powerful CL baseline. In addition, we find that for complex datasets, when memory size is larger than 5000 (10\% of the training data), most of the replay-based methods (except for GSS) outperform the iid-online, a baseline that trains the model with a one-pass through the data but with iid-sampled mini-batch from the whole dataset. This means that storing a small number of training samples is crucial for combating forgetting as well as the learning of the current task in the online class-incremental setting. 

We also verify claims from previous work \cite{lesort2019regularization, farquhar2018robust, mir}. EWC, a regularization-based method, not only is surpassed by all memory-based methods but also underperforms the fine-tuning baseline. Additionally,  AGEM, a method that uses memory samples to constrain parameter updates,  delivers worse performance compared with reply-based methods (ER, MIR, and GSS), especially when memory size increases. 

Overall, by evaluating on three standard CL datasets and comparing to the state-of-the-art CL methods, we have shown the effectiveness of ASER and its variant ASER$_\mu$ in overcoming catastrophic forgetting, especially in more complex datasets and memory size is relatively small.

\section{Conclusion} \label{sec:conclusion}

In this work, we proposed a novel ASER method that scores memory data samples according to
their ability to preserve latent decision boundaries for previously observed classes while interfering with latent decision boundaries of current classes being learned. Overall, in the online class-incremental setting, we observed that ASER and its ASER$_\mu$ variant provide competitive or improved performance on a variety of datasets compared to state-of-the-art ER-based continual learning methods. We also remark that this work paves the way for a number of interesting research directions building on this work.
Although our SV-based method has greatly improved the memory retrieval and update strategies, we may be able to do better than simply concatenating retrieved samples with the incoming batch.  Hence, future work could focus on more sophisticated methods to utilize the retrieved samples. 
It would also be interesting to investigate alternate CL-specific utility function variations for SV.

\section*{Acknowledgements}
This research was supported by LG AI Research.

\bibliography{references.bib}

\appendix
\counterwithin{figure}{section}
\counterwithin{table}{section}

\begin{center}\LARGE \textbf{Appendix}\end{center}

\normalsize

\section{Properties of KNN Shapley Value} \label{appendix:knnsv}
\setcounter{equation}{0}
\counterwithin{equation}{section}

We prove the properties of KNN-SV presented in Section \ref{sec:aser}. Given a candidate sample $i=(\mathbf{x}_i, y_i)\in D_c$ and an evaluation point $j=(\mathbf{x}_j^\text{ev},y_j^\text{ev})$, where $D_c$ is a candidate set with $|D_c| =N_c$, we denote the KNN-SV of the point $i$ w.r.t. the evaluation point $j$ as $s_j(i)$. For notational convenience, we assume that points in $D_c$ are sorted based on distances from the evaluation point in ascending order. In other words, $d(i, j) \le d(i', j)~~\forall i < i'$ where $d(i,j)$ is the distance between $\mathbf{x}_i$ and $\mathbf{x}_j^\text{ev}$.

\begin{property}
$s_j(i) > 0$ if and only if $y_i = y^\text{ev}_j$. Also, $s_j(i)=0$ only when $S=\{i'|y_{i'}=y^\text{ev}_j, ~\forall i'\in \{i+1,\dots,N_c\}\}=\emptyset$.
\end{property}
\begin{proof}
Firstly, we prove $s_j(i) > 0$ \textit{if and only if} $y_i = y^\text{ev}_j$ along with another statement, $|s_j(i)|<\frac{1}{i-1}$. The proof is by induction, starting from the base case when $i=N_c$. When $i=N_c$, $s_j(N_c)=\frac{\mathds{1}(y_{N_c}=y_j^\text{ev})}{N_c}$ as per \eqref{eq:farthestsv}. Hence, $s_j(N_c)>0$ holds iff $y_{N_c}=y_j^\text{ev}$. Additionally, we see that $|s_j(N_c)|\le \frac{1}{N_c}<\frac{1}{N_c -1}$. We can also verify the case for $i=N_c-1$ using \eqref{eq:recur_knnsv}:
\small
\begin{align}
    s_j(N_c-1)=\begin{cases}
        \frac{1}{N_c}>0 & \text{if}~~y_{N_c-1}=y_{N_c}=y_j^\text{ev}\\
        \frac{1}{N_c -1}>0 & \text{if}~~y_{N_c -1}=y_j^\text{ev}\ne y_{N_c}\\
        \frac{1}{N_c} - \frac{1}{N_c-1}<0 &\text{if}~~y_{N_c-1}\ne y_j^\text{ev}~~\text{and} \\ & \quad y_{N_c} = y_j^\text{ev}\\
        0 & \text{if}~~y_{N_c-1}\ne y_j^\text{ev} ~~\text{and}\\
        &\quad y_{N_c}\ne y_j^\text{ev}
    \end{cases}
\end{align}
\normalsize
So, we again note that $s_j(N_c-1)>0$ iff $y_{N_c-1}=y_j^\text{ev}$, and $|s_j(N_c-1)|\le \frac{1}{N_c-1}<\frac{1}{N_c-2}$ holds. 

Now, assume for $i=m\ge K+1$, $s_j(m)>0$ iff $y_m=y_j^\text{ev}$ and $|s_j(m)|<\frac{1}{m-1}$. Then, for $i=m-1\ge K$,
\small
\begin{align}
    s_j(m-1)&=s_j(m)+\frac{\mathds{1}(y_{m-1}=y_j^\text{ev})-\mathds{1}(y_m=y_j^\text{ev})}{m-1}\nonumber\\
    &=\begin{cases}
    s_j(m)>0 & \text{if}~~y_{m-1}=y_m=y_j^\text{ev},\\
    s_j(m) + \frac{1}{m-1} > 0& \text{if}~~y_{m-1}=y_j^\text{ev}\ne y_m,\\
    s_j(m)-\frac{1}{m-1}<0 & \text{if}~~y_{m-1}\ne y_j^\text{ev} ~~\text{and}\\
    & \quad y_m=y_j^\text{ev},\\
    s_j(m)\le 0 & \text{if}~~y_{m-1}\ne y_j^\text{ev}~~\text{and}\\
    &\quad y_{m}\ne y_j^\text{ev}
    \end{cases} \label{eq:cases}
\end{align}
\normalsize
Note that the second and the third cases in \eqref{eq:cases} hold because $|s_j(m)|<\frac{1}{m-1}$ by assumption. Additionally, it is straightforward to check that $|s_j(m-1)|<\frac{1}{m-2}$ holds for all cases.  Hence, we have shown the statement holds for $m\ge K$. 

For $m<K$, we firstly note that $|s_j(K)|\le \nicefrac{1}{K}$ because $|s_j(K+1)|<\nicefrac{1}{K}$ and $s_j(K)=s_j(K+1)+\frac{\mathds{1}(y_K=y_j^\text{ev})-\mathds{1}(y_{K+1}=y_j^\text{ev})}{K}$. Then, we see that the increment and the decrement, if any, are always $\frac{1}{K}$ for $m<K$. Therefore, $s_j(m)>0$ iff $y_m=y_j^\text{ev}$. Furthermore, we have noted that the sign of $s_j(i)$ has to change iff $\mathds{1}(y_i=y_j^\text{ev})\ne \mathds{1}(y_{i+1}=y_j^\text{ev})$ because $|s_j(i+1)|<\nicefrac{1}{i}$. This implies $s_j(i)=0$ only when $S=\{i'|y_{i'}=y^\text{ev}_j, ~\forall i'\in \{i+1,\dots,N_c\}\}=\emptyset$, and we conclude the proof.
\end{proof}

\begin{property}
$|s_j(m)|$ is a non-increasing function of $m$ for $m$ such that $y_m=y_j^\text{ev}$. Similarly, $|s_j(n)|$ is a non-increasing function of $n$ for $n$ such that $y_n\ne y_j^\text{ev}$. And for $m\ge K$, $|s_j(m)|-|s_j(m')|>0$ holds for $m<m'$, where $m'$ is the smallest index with $\mathds{1}(y_m=y_j^\text{ev})=\mathds{1}(y_{m'}=y_j^\text{ev})$, if there exists $l\in (m, m')$ such that $\mathds{1}(y_l=y_j^\text{ev})\ne\mathds{1}(y_{m}=y_j^\text{ev})$. In other words, as $i$ gets closer to the evaluation point $j$, $|s_j(i)|$ cannot decrease for points with the same $\mathds{1}(y_i=y_j^\text{ev})$, and for $i\ge K$, it can only increase when there exist more than one differently labeled points.
\end{property}
\begin{proof}
We only show for $m$ such that $y_m=y_j^\text{ev}$ as it can be similarly done for $n$ with $y_n\ne y_j^\text{ev}$. If $y_l=y_j^\text{ev}~$ $\forall l\in(m,m']$, then it trivially holds that $s_j(m)=s_j(m')$. Now, assume that there exists one $l\in (m,m')$ such that $y_l\ne y_j^\text{ev}$ and that $l=m+1$. Then, we see that $|s_j(m)|-|s_j(m')|>0$ holds. This is because
\small
\begin{align}
    s_j(m)&=s_j(m+1)+\frac{\mathds{1}(y_m=y_j^\text{ev})-\mathds{1}(y_{m+1}=y_j^\text{ev})}{m} \nonumber\\
    &=s_j(m+2)+\frac{\mathds{1}(y_{m+1}=y_j^\text{ev})-\mathds{1}(y_{m+2}=y_j^\text{ev})}{m+1} + \frac{1}{m}\nonumber\\
    &= s_j(m+2) - \frac{1}{m+1} + \frac{1}{m} \nonumber\\
    &= s_j(m+2) + \big( \frac{1}{m}-\frac{1}{m+1}\big)\nonumber\\
    &>s_j(m+2)=s_j(m'),~~\forall m'\ge m+2 \label{eq:B2}
\end{align}
\normalsize
Then, we note that $s_j(m'')=s_j(m)$ for all $m''\in (i, m]$ where $i<m$ is the largest index with $y_i\ne y_j^\text{ev}$ (if exists) or $i=0$. This shows that $|s_j(m)|>|s_j(m')|$ holds for $m<m'$ if there is a single $l\in(m,m')$ with $y_l \ne y_j^\text{ev}$. When there are multiple (possibly consecutive) points with $y_l\ne y_j^\text{ev}$, we can always select $\hat{m}\ge m$ such that there is only a single point (or several consecutive points) $l$ with $y_l\ne y_j^\text{ev}$, leading to $|s_j(\hat{m})|>|s_j(m')|$. By applying this multiple times, we get $|s_j(m)|>s_j(m')|$. 
\end{proof}

{\setlength\tabcolsep{2.1pt}
\begin{table*}
\begin{subtable}{.365\textwidth}
    \centering
    \small
    \begin{tabular}{c c c c} 
    \toprule
    Method &M=1k& M=2k&  M=5k\Bstrut \\
    \hline\hline
    \Tstrut
    iid online & $14.7\pm0.6$&$14.7\pm0.6$&$14.7\pm0.6$\\
    iid offline & $42.4\pm0.4$&$42.4\pm0.4$&$42.4\pm0.4$\\
    \midrule
    AGEM & $7.0\pm0.4$ &$7.1\pm0.5$ &$6.9\pm0.7$ \\
    ER & $8.7\pm0.4$&$11.8\pm0.9$&$16.5\pm0.9$\\
    EWC & $3.1\pm0.3$ &$3.1\pm0.3$ &$3.1\pm0.3$\\
    fine-tune&$4.3\pm0.2$&$4.3\pm0.2$&$4.3\pm0.2$\\
    GSS&$7.5\pm0.5$&$10.7\pm0.8$&$12.5\pm0.4$\\
    MIR&$8.1\pm0.3$&$11.2\pm0.7$&$15.9\pm1.6$\\
    \midrule
    ASER&$\mathbf{11.7\pm0.8}$&$\mathbf{14.4\pm0.4}$&$\mathbf{18.2\pm0.7}$\\
    ASER$_\mu$&$\mathbf{12.2\pm0.8}$&$\mathbf{14.8\pm1.1}$&$\mathbf{18.2\pm1.1}$\\
    Dist&$7.9\pm0.5$&$10.4\pm0.7$&$15.5\pm0.9$\\
    Dist$_\mu$&$8.1\pm0.5$&$9.1\pm0.7$&$14.9\pm0.9$\\
    \hline
    \bottomrule
    \end{tabular}
    \subcaption{Mini-ImageNet}
\end{subtable}
\hfill \setlength\tabcolsep{2.1pt}
\begin{subtable}{.295\textwidth}
    \centering
\small
\begin{tabular}{c c c } 
    \toprule
     M=1k&  M=2k &  M=5k \Bstrut \\   
    \hline\hline
    \Tstrut
    $20.5\pm0.4$&$20.5\pm0.4$&$20.5\pm0.4$\\
    $47.4\pm0.3$&$47.4\pm0.3$&$47.4\pm0.3$\\
    \midrule
    $9.5\pm0.4$&$9.3\pm0.4$&$9.7\pm0.3$\\
    $11.2\pm0.4$&$14.6\pm0.4$&$20.1\pm0.8$\\
    $4.8\pm0.2$&$4.8\pm0.2$&$4.8\pm0.2$\\
    $5.9\pm0.2$&$5.9\pm0.2$&$5.9\pm0.2$\\
    $9.3\pm0.2$&$10.9\pm0.3$&$15.9\pm0.4$\\
    $11.2\pm0.3$&$14.1\pm0.2$&$21.2\pm0.6$\\
    \midrule
    $12.3\pm0.4$&$14.7\pm0.7$&$20.0\pm0.6$\\
    $\mathbf{14.0\pm0.4}$&$\mathbf{17.2\pm0.5}$&$21.7\pm0.5$\\
    $10.3\pm.3$&$12.4\pm0.5$&$16.7\pm0.6$\\
    $10.5\pm0.2$&$13.4\pm0.4$&$17.2\pm0.8$\\
    \hline
    \bottomrule
\end{tabular}
\subcaption{CIFAR-100}
\end{subtable}\hfill
\begin{subtable}{.295\textwidth}
    \centering
\small
\begin{tabular}{c c c} 
    \toprule
    M=0.2k &  M=0.5k & M=1k \Bstrut \\ 
    \hline\hline
    \Tstrut
    $62.9\pm1.5$&$62.9\pm1.5$&$62.9\pm1.5$ \\
    $79.7\pm0.4$&$79.7\pm0.4$&$79.7\pm0.4$ \\
    \midrule
    $22.7\pm1.8$&$22.7\pm1.9$&$22.6\pm0.7$\\
    $26.4\pm1.0$&$32.2\pm1.4$&$38.4\pm1.7$\\
    $17.9\pm0.3$&$17.9\pm0.3$&$17.9\pm0.3$\\
    $17.9\pm0.4$&$17.9\pm0.4$&$17.9\pm0.4$ \\
    $26.9\pm1.2$&$30.7\pm1.2$&$40.1\pm1.4$\\
    $28.3\pm1.6$&$35.6\pm1.2$&$42.4\pm1.5$\\
    \midrule
    $27.8\pm1.0$&$36.2\pm1.1$&$43.1\pm1.2$\\
    $26.4\pm1.5$&$36.3\pm1.2$&$43.5\pm1.4$\\
    $22.9\pm0.9$&$31.6\pm1.7$&$38.0\pm2.3$\\
    $23.4\pm1.0$&$29.7\pm1.2$&$35.6\pm1.3$\\
    \hline
    \bottomrule
\end{tabular}
\subcaption{CIFAR-10}
\end{subtable} 
\caption{Average Accuracy(higher is better). Memory buffer size M.}
\label{tab:acc_more}
\end{table*}
}

{\setlength\tabcolsep{2.1pt}
\begin{table*}
\begin{subtable}{.365\linewidth}
    \centering
    \label{tab:forget_mainting_miniimagenet}
    \small
    \begin{tabular}{ c c c c} 
    \toprule
    Method &M=1k &M=2k &  M=5k\Bstrut \\ 
    \hline\hline
    \Tstrut
    AGEM & $29.3\pm0.9$ &$30.0\pm0.9$ &$29.9\pm0.8$ \\
    ER & $29.7\pm1.3$&$29.2\pm0.9$&$26.6\pm1.1$\\
    EWC & $28.1\pm0.8$ &$28.1\pm0.8$ &$28.1\pm0.8$ \\
    fine-tune &$35.6\pm0.9$&$35.6\pm0.9$&$35.6\pm0.9$\\
    GSS &$29.6\pm1.2$& $27.4\pm1.1$&$29.9\pm1.2$\\
    MIR &$29.7\pm1.0$& $27.2\pm1.1$&$26.2\pm1.4$\\
    \midrule
    ASER &$30.1\pm1.3$& $24.7\pm1.0$&$20.9\pm1.2$\\
    ASER$_\mu$ &$28.0\pm1.3$&$\mathbf{22.2\pm1.6}$&$\mathbf{17.2\pm1.4}$\\
    Dist&$30.4\pm1.1$&$25.5\pm1.7$&$26.3\pm1.2$\\
    Dist$_\mu$&$29.7\pm1.1$&$29.8\pm1.1$&$27.1\pm1.4$\\
    \hline
    \bottomrule
    \end{tabular}
    \subcaption{Mini-ImageNet}
\end{subtable}
\hfill
\begin{subtable}{.295\linewidth}
    \centering
\small
\begin{tabular}{c c c } 
    \toprule
    M=1k& M=2k &  M=5k\Bstrut \\ 
    \hline\hline
    \Tstrut
    $40.4\pm0.7$&$39.7\pm0.8$&$39.8\pm1.0$\\
    $45.0\pm0.5$&$40.5\pm0.8$&$34.5\pm0.8$\\
    $39.1\pm1.2$&$39.1\pm1.2$&$39.1\pm1.2$\\
    $50.4\pm1.0$&$50.4\pm1.0$&$50.4\pm1.0$\\
    $46.9\pm0.7$&$42.3\pm0.8$&$39.2\pm0.9$\\
    $45.5\pm0.8$&$40.4\pm0.6$&$31.4\pm0.6$\\
    \midrule
    $50.1\pm0.6$&$45.9\pm0.9$&$36.7\pm0.8$\\
    $45.0\pm0.7$&$\mathbf{38.6\pm0.6}$&$\mathbf{30.3\pm0.5}$\\
    $48.1\pm0.8$&$43.8\pm0.5$&$39.5\pm0.8$\\
    $48.0\pm0.6$&$43.5\pm0.6$&$39.9\pm0.8$\\
    \hline
\bottomrule
\end{tabular}
\subcaption{CIFAR-100}
\end{subtable}\hfill
\begin{subtable}{.295\linewidth}
        \centering
    \small
    \begin{tabular}{c c c } 
        \toprule
        M=0.2k &  M=0.5k & M=1k \Bstrut\\ 
        \hline\hline
        \Tstrut
        $36.1\pm3.8$&$43.2\pm4.2$&$48.1\pm3.0$\\
        $72.8\pm1.7$&$63.1\pm2.4$&$55.8\pm2.6$\\
        $81.5\pm1.4$&$81.5\pm1.4$&$81.5\pm1.4$\\
        $81.7\pm0.7$&$81.7\pm0.7$&$81.7\pm0.7$ \\
        $75.5\pm1.5$&$65.9\pm1.6$&$54.9\pm2.0$\\
        $\mathbf{67.0\pm2.6}$&$68.9\pm1.7$&$\mathbf{47.7\pm2.9}$\\
        \midrule
        $71.1\pm1.8$&$59.1\pm1.5$&$50.4\pm1.5$\\
        $72.4\pm1.9$&$58.8\pm1.4$&$47.9\pm1.6$\\
        $76.4\pm2.3$&$\mathbf{63.7\pm2.3}$&$53.6\pm3.8$\\
        $77.6\pm1.6$&$68.6\pm2.0$&$58.8\pm2.7$\\
        \hline
        \bottomrule
    \end{tabular}
    \subcaption{CIFAR-10}
    \end{subtable}
\caption{Average Forgetting (lower is better). Memory buffer size M.}
\label{tab:forget_more}
\end{table*}
}

\section{Detailed Performance Evaluation} \label{appendix: performance}

Following the definition of \textbf{ASV} and \textbf{ASV$_\mu$}, we can replace the Shapley value with distance (we use Euclidean as example) in \eqref{eq:adv_obj_minmax} and \eqref{eq:adv_obj_mean}. Specifically, we want to retrieve a point $i$ such that its distances from samples of the same label in $\mathcal{M}$ and its distances from input batch samples are both small. Concretely, the score for a candidate point $i$ is defined as follows:
\begin{align}
    \textbf{Dist}(i) = - \Big[ \min_{j\in S_\text{sub}(i)} d(i, j) + \min_{k \in B_n} d(i, k)\Big],
    \label{eq:dist_minmax}
\end{align}
as well as a ``softer'' mean variation \textbf{Dist}$_\mu$
\begin{align}
    \textbf{Dist}_\mu(i) = - \bigg[ \frac{1}{|S_\text{sub}(i)|}{\sum_{j\in S_\text{sub}(i)} d(i, j)} + \frac{1}{b}{\sum_{k\in B_n} d(i, k)}\bigg].
    \label{eq:dist_mean}
\end{align}
Here, $i \in \mathcal{M} \setminus S_\text{sub}$ and $S_\text{sub}$ is defined as in Section \ref{sec:aser} and $S_\text{sub}(i)=\{i'|i'\in S_\text{sub}~~ \&~~ y_{i'}=y_i\}$. $d(i, j)$ is the Euclidean distance between the candidate point $i$ and an evaluation point $j$ in the latent space. Finally, we simply replace the score in Algorithm \ref{alg:algo_aser} (line 9) with either one of the above scores.

\subsection{Detailed Result Tables}
In addition to the algorithms listed in Section~\ref{sec:experiment}, in Table~\ref{tab:acc_more} and Table~\ref{tab:forget_more}, we include more baselines for comparison:
\begin{itemize}
  \item\textbf{AGEM} \cite{agem}: Averaged Gradient Episodic Memory, a memory-based method that utilizes the samples in the memory buffer to constrain the parameter updates.  
  \item \textbf{ASER \& ASER$_\mu$}: Our proposed methods. ASER scores samples in the memory with \textbf{ASV} in \eqref{eq:adv_obj_minmax}. ASER$_\mu$ uses the mean variation \textbf{ASV}$_\mu$ in \eqref{eq:adv_obj_mean}.
  \item \textbf{Dist \& Dist$_\mu$}: The Euclidean variants of ASER \& ASER$_\mu$ that replace Shapley value with Euclidean distance, as described above.
  \item \textbf{ER} \cite{tiny}: Experience replay, a recent and successful rehearsal method with random sampling in \retrieval{} and reservoir sampling in \update{}.
  \item \textbf{EWC} \cite{Kirkpatrick2017}: Elastic Weight Consolidation, a prior-focused method that limits the update of parameters that were important to the past tasks, as measured by the Fisher information matrix. 
  \item \textbf{GSS} \cite{gss}: Gradient-Based Sample Selection, a \update{}  method that diversifies the gradients of the samples in the replay memory.
  \item \textbf{MIR} \cite{mir}: Maximally Interfered Retrieval, a \retrieval{} method that retrieves memory samples that suffer from an increase in loss given the estimated parameters update based on the current task.
  \item \textbf{iid-online \& iid-offline}: iid-online trains the model with a single-pass through the same set of data, but each mini-batch is sampled iid from the training set. iid-offline trains the model over multiple epochs on the dataset with iid sampled mini-batch. We use 5 epochs for iid-offline in all the experiments as in \cite{mir, gss}. 
  \item \textbf{fine-tune}: As an important baseline in previous work \cite{mir, gss, dirichlet}, it simply trains the model in the order the data is presented without any specific method for forgetting avoidance.
\end{itemize}

\subsubsection{Average Accuracy and Forgetting}
As we can see from Table~\ref{tab:acc_more} and Table~\ref{tab:forget_more}, ASER and ASER$_\mu$ outperform Dist and Dist$_\mu$. The reason may be that both ASER methods do not greedily retrieve samples with the smallest distances to either $S_\text{sub}$ (sub-sample from $\mathcal{M}$) or $B_n$(incoming mini-batch). This is because for a single evaluation point $j$, $s_j(\alpha_m)=s_j(\alpha_{m+1})$ when $y_{\alpha_m}=y_{\alpha_{m+1}}$. So, a few points can have the same score even if some of them are farther from the evaluation point. 

We also verify some claims from previous work \cite{lesort2019regularization, farquhar2018robust, mir}. EWC, a prior-focused method, not only is surpassed by all memory-based methods but also underperforms the fine-tuning baseline. Additionally,  AGEM, a method that uses memory samples to constrain parameter updates,  delivers worse performance compared with reply-based methods (ER, MIR, and GSS), especially when memory size increases. 

\subsubsection{Training Time}
\begin{figure}
    \centering
    \includegraphics[
    height=5cm]{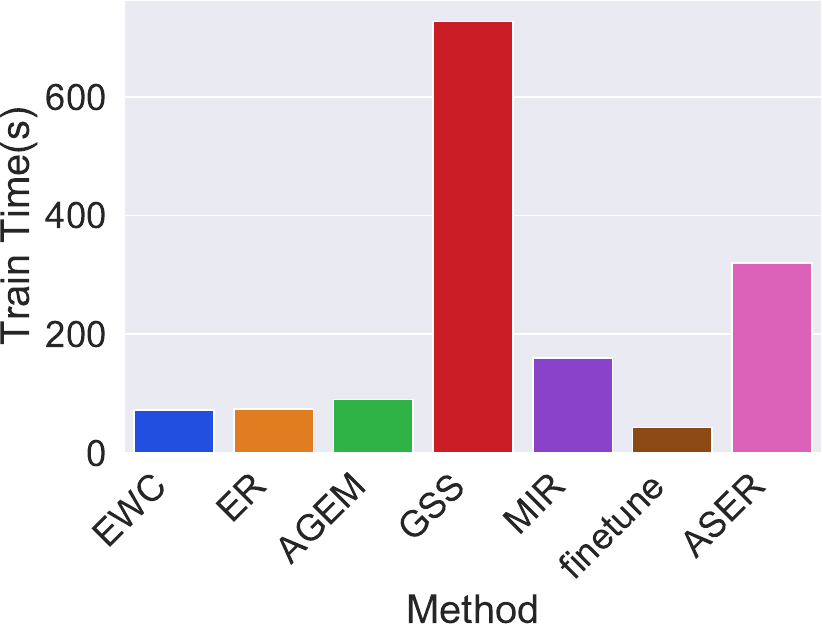}

    \caption{Training time comparison for CIFAR-100}\label{fig:train_time}
\end{figure}

EWC, ER and AGEM have similar training time and their training times are almost twice as the finetune baseline. Since MIR, GSS and our proposed ASER need to perform additional calculation during \retrieval{} and \update{}, the training times are longer than the methods mentioned above. ASER takes longer than MIR because MIR only has additional computation in \retrieval{} but ASER carefully selects samples in both \retrieval{} and \update{}. Compared with GSS, ASER is more computationally efficient. 
Figure.\ref{fig:train_time}

\section{Ablation Studies} \label{appendix:ablation}
{\setlength\tabcolsep{2.0pt}
\begin{table*}[h!]
\begin{subtable}{.37\textwidth}
    \centering
    \label{tab:acc_ablation_mini}
    \small
    \begin{tabular}{c c c c} 
    \toprule
    Method &M=1k & M=2k &  M=5k\Bstrut \\
    \hline\hline
        \Tstrut
    ER & $8.7\pm0.4$&$11.8\pm0.9$&$16.5\pm0.9$\\
    SV-upd & $\mathbf{13.4\pm0.8}$&$\mathbf{15.5\pm0.7}$&$\mathbf{18.4\pm0.4}$\\
    ASV-ret & $6.9\pm0.4$ &$9.9\pm0.9$ &$16.2\pm0.8$ \\
    ASV$_\mu$-ret & $7.4\pm0.6$ &$10.0\pm1.0$ &$17.1\pm0.9$ \\
    ASER&$11.7\pm0.8$&$14.4\pm0.4$&$\mathbf{18.2\pm0.7}$\\
    ASER$_\mu$&$12.2\pm0.8$&$\mathbf{14.8\pm1.1}$&$\mathbf{18.2\pm1.1}$\\
    \hline
    \bottomrule
    \end{tabular}
    \subcaption{Mini-ImageNet}
\end{subtable}
\hfill \setlength\tabcolsep{2.0pt}
\begin{subtable}{.29\textwidth}
    \centering
    \label{tab:acc_ablation_cifar100}
\small
\begin{tabular}{c c c } 
    \toprule
     M=1k&  M=2k &  M=5k \Bstrut \\   
    \hline\hline
    \Tstrut
    $11.2\pm0.4$&$14.6\pm0.4$&$20.1\pm0.8$\\
    $\mathbf{14.0\pm0.6}$&$\mathbf{17.2\pm0.4}$&$20.9\pm0.6$\\
    $10.1\pm0.3$&$13.9\pm0.3$&$20.3\pm0.3$\\
    $10.8\pm0.3$&$14.8\pm0.4$&$\mathbf{21.7\pm0.3}$\\
    $12.3\pm0.4$&$14.7\pm0.7$&$20.0\pm0.6$\\
    $\mathbf{14.0\pm0.4}$&$\mathbf{17.2\pm0.5}$&$\mathbf{21.7\pm0.5}$\\
    \hline
    \bottomrule
\end{tabular}
\subcaption{CIFAR-100}
\end{subtable}\hfill \setlength\tabcolsep{2.0pt}
\begin{subtable}{.29\textwidth}
    \centering
    \label{tab:acc_ablation_cifar10}
\small
\begin{tabular}{c c c} 
    \toprule
    M=0.2k &  M=0.5k & M=1k \Bstrut \\ 
    \hline\hline
    \Tstrut
    $26.4\pm1.0$&$32.2\pm1.4$&$38.4\pm1.7$ \\
    $25.9\pm0.6$&$33.7\pm1.4$&$41.2\pm1.3$ \\
    $26.6\pm1.0$&$34.7\pm1.1$&$39.3\pm1.6$ \\
    $25.8\pm1.0$&$35.7\pm1.4$&$40.2\pm1.0$\\
    $\mathbf{27.8\pm1.0}$&$36.2\pm1.1$&$\mathbf{43.1\pm1.2}$\\
    $26.4\pm1.5$&$36.3\pm1.2$&$\mathbf{43.5\pm1.4}$\\
    \hline
    \bottomrule
\end{tabular}
\subcaption{CIFAR-10}
\end{subtable} 
\caption{Ablation analysis. Average Accuracy (higher is better). Memory buffer size M.}
\label{tab:acc_ablation}
\end{table*}
}

{\setlength\tabcolsep{2.0pt}
\begin{table*}[h!]
\begin{subtable}{.3775\textwidth}
    \centering
    \label{tab:forgetting_mini_ablation}
    \small
    \begin{tabular}{c c c c} 
    \toprule
    Method &M=1k & M=2k &  M=5k\Bstrut \\
    \hline\hline
        \Tstrut
    ER & $29.7\pm1.3$&$29.2\pm0.9$&$26.6\pm1.1$\\
    SV-upd & $29.2\pm1.1$&$26.4\pm1.5$&$24.6\pm1.3$\\
    ASV-ret & $\mathbf{27.6\pm1.2}$ &$27.7\pm1.3$ &$25.5\pm1.2$ \\
    ASV$_\mu$-ret & $\mathbf{28.3\pm1.3}$ &$27.4\pm1.2$ &$23.2\pm1.6$ \\
    ASER&$30.1\pm1.3$&$24.7\pm1.0$&$20.9\pm1.2$\\
    ASER$_\mu$&$\mathbf{28.0\pm1.3}$&$\mathbf{22.2\pm1.6}$&$\mathbf{17.2\pm1.4}$\\
    \hline
    \bottomrule
    \end{tabular}
    \subcaption{Mini-ImageNet}
\end{subtable}
\hfill \setlength\tabcolsep{2.0pt}
\begin{subtable}{.29\textwidth}
    \centering
\small
\begin{tabular}{c c c } 
    \toprule
     M=1k&  M=2k &  M=5k \Bstrut \\   
    \hline\hline
    \Tstrut
    $45.5\pm0.5$&$40.5\pm0.8$&$34.5\pm0.8$\\
    $45.9\pm0.7$&$41.6\pm0.7$&$36.6\pm0.5$\\
    $47.4\pm0.4$&$42.0\pm0.6$&$34.5\pm0.8$\\
    $45.5\pm0.7$&$\mathbf{37.4\pm0.5}$&$31.8\pm0.6$\\
    $50.1\pm0.6$&$45.9\pm0.9$&$36.7\pm0.8$\\
    $45.0\pm0.7$&$38.6\pm0.6$&$\mathbf{30.3\pm0.5}$\\
    \hline
    \bottomrule
\end{tabular}
\subcaption{CIFAR-100}
\end{subtable}
\hfill \setlength\tabcolsep{2.0pt}
\begin{subtable}{.29\textwidth}
    \centering
\small
\begin{tabular}{c c c} 
    \toprule
    M=0.2k &  M=0.5k & M=1k \Bstrut \\ 
    \hline\hline
    \Tstrut
    $72.8\pm1.7$&$63.1\pm2.4$&$55.8\pm2.6$ \\
    $73.6\pm1.0$&$63.4\pm1.6$&$52.4\pm2.0$ \\
    $71.7\pm2.1$&$60.1\pm1.7$&$53.5\pm2.5$ \\
    $73.6\pm1.4$&$58.3\pm2.2$&$49.0\pm2.2$\\
    $71.1\pm1.8$&$59.1\pm1.5$&$50.4\pm1.5$\\
    $72.4\pm1.9$&$58.8\pm1.4$&$\mathbf{47.9\pm1.6}$\\
    \hline
    \bottomrule
\end{tabular}
\subcaption{CIFAR-10}
\end{subtable} 
\caption{Ablation analysis. Average Forgetting (lower is better). Memory buffer size M.}
\label{tab:ablation_forgetting}
\end{table*}
}
In ASER \& ASER$_\mu$, we use ASV and ASV$_\mu$ for scoring samples for \retrieval{}, while KNN-SV is used for scoring samples for \update{}. In this part, we examine several ablations to understand contributions of each component in ASER methods. In addition to ASER, ASER$_\mu$ and ER as in Section \ref{sec:experiment}, we compare 3 more variations:
\begin{itemize}
    \item \textbf{SV-upd}: Use KNN-SV \update{} as described in Section \ref{sec:aser} while randomly retrieving samples from the memory for replay.
    \item \textbf{ASV-ret}: Use \eqref{eq:adv_obj_minmax} scoring function for \retrieval{} while using reservoir sampling for \update{}.
    \item \textbf{ASV}$_\mu$\textbf{-ret}: Use \eqref{eq:adv_obj_mean} scoring function for \retrieval{} while using reservoir sampling for \update{}.
\end{itemize}
Table \ref{tab:acc_ablation} compares the average accuracy of these variations. As for CIFAR-10, we can see that all SV-based methods improve upon ER. In particular, ASER and ASER$_\mu$ show the largest improvements, suggesting the effectiveness of the combination of the KNN-SV based \retrieval{} and \update{}. For the other two datasets, it turns out that SV-upd is a powerful \update{} method. Compared to GSS \cite{gss} which suggests another \update{} method, we observe significant performance boosts (see Table \ref{tab:acc_main}). In these two datasets, ASV($_\mu$)-ret methods and ER perform comparably. However, we note that we can further fight the forgetting when \retrieval{} and \update{} are used together (Table \ref{tab:ablation_forgetting}). In summary, ASER with KNN-SV \update{} performs competitively or better than the variations, underscoring the importance of SV-based methods for both \update{} and \retrieval{}.

\section{Dataset Detail}\label{appendix: dataset detail}
\begin{table*}
    \small
    \centering
\begin{tabular}{l | c c c} 
    
    \toprule
    &Split Mini-ImageNet &  Split CIFAR-100 & Split CIFAR-10 \Bstrut \\ 
    \hline\hline
    \Tstrut
    num. of tasks& 10&10&5\\
    image size& 3x84x84& 3x32x32&3x32x32\\
    num. of classes per task&10&10&2\\
    num. of training images per task&4800&4500&9000\\
    num. of validation images per task&600&500&1000\\
    num. of testing images per task&600&1000&1000\\
    \hline
    \bottomrule
\end{tabular}
\vspace{0.3cm}
\caption{Dataset statistics}
\label{tab:dataset}
\end{table*}

Table~\ref{tab:dataset} shows the summary of the datasets used for the experiments. For a fair comparison, the classes in each task and the order of tasks are fixed in all experiments. For Split CIFAR-10, the first task contains class [0, 1], the second task contains class [2, 3], and so on. For Split CIFAR-100, similar to Split CIFAR-10, the first task contains class [0, 1, \dots, 9], the second task contains class [10, 11, \dots, 19] and so on. 

In original Mini-ImageNet, 100 classes are divided into 64, 16, and 20 classes respectively for meta-training, meta-validation, and meta-test \cite{miniimagenet}. For Split Mini-ImageNet, we firstly combine 64, 16, and 20 classes into one dataset. The first task contains the first 10 classes; the second task contains the next 10 classes, and so on. 

\section{Detail of Experiments} \label{appendix:tuning}
We use a reduced ResNet18, similar to \cite{tiny, gem}, as the base model for all datasets, and the network is trained via cross-entropy loss with SGD optimizer. Note that several replay-based SOTA continual learning algorithms have also used the simple SGD \cite{mir, gss, agem, tiny, gem}. For all experiments, we use the learning rate of 0.1 following the same setting as in \citet{mir}. The mini-batch size is 10 and the size of the mini-batch retrieved from memory is also set to 10 irrespective of the size of the memory. Since we apply the online setting, the model only sees each batch once, so the number of epochs is set to 1 for all experiments. 

We have used the memory size $M=$ 1k, 2k and 5k for Mini-ImageNet and CIFAR-100, while $M=$ 0.2k, 0.5k and 1k for CIFAR-10. As for CIFAR-10, we use the same memory sizes as in MIR \cite{mir} (0.4\%, 1.1\% and 2.2\%); however, we found that they used disproportionately bigger memory for Mini-ImageNet (20\% of the training data). One of the key desiderata of continual learning for deployment is limited memory footprint \cite{agem, tiny, farquhar2018robust, Parisi2019}. Hence, we instead use smaller sizes of memory for both CIFAR-100 and Mini-ImageNet (around 2\%, 4\% and 10\% of the training data) that better reflect real-world use cases with a high ratio of data to memory.

As for the hyperparameters of baselines, we tune the number of samples used for computing maximal gradients cosine similarity for GSS. For MIR, we tune the number of subsamples used to apply the MIR search criterion using the validation sets.

As for the hyperparameters of baselines, we use the validation sets to tune the number of samples (S) used for computing maximal gradients cosine similarity for GSS; the number of subsamples (C) used to apply the MIR search criterion for MIR. We have tuned two hyperparameters for ASER: the number of candidate samples ($N_c$) and the number of neighbors ($K$) for KNN-SV computation. Details of the datasets used in the experiment are shown in Table \ref{tab:dataset} in Appendix \ref{appendix: dataset detail}. We have summarized the hyperparameters used in the experiments in Table \ref{tab:hyperparameters}.

The code to reproduce all results can be found in the attached zip file. 

{
\setlength\tabcolsep{3.5pt}
\begin{table*}[ht!]
\centering
\begin{subtable}{0.26\textwidth}
    \centering
    \small
    \begin{tabular}{c c c} 
    Method & & \Bstrut \\
    \hline\hline
    \Tstrut
    GSS & \multicolumn{2}{c}{S=20} \\
    MIR & \multicolumn{2}{c}{C=100} \\
    ASER &  $K=7$ & $N_c=200$ \\
    ASER$_\mu$& $K=3$ & $N_c=250$ \Bstrut\\
    \bottomrule
    \end{tabular}
    \subcaption{Mini-ImageNet}
\end{subtable}
\begin{subtable}{.165\textwidth}
    \centering
\small
\begin{tabular}{c c} 
    \multicolumn{2}{c}{Hyperparameters} \Bstrut \\   
    \hline\hline
    \multicolumn{2}{c}{S=10} \Tstrut\\
    \multicolumn{2}{c}{C=50} \\
    $K=1$ & $N_c=350$ \\
    $K=3$ & $N_c=150$ \Bstrut \\
    \bottomrule
\end{tabular}
\subcaption{CIFAR-100}
\end{subtable}
\begin{subtable}{.165\textwidth}
    \centering
\small
\begin{tabular}{c c} 
    \Bstrut \\   
    \hline\hline
    \multicolumn{2}{c}{S=10} \Tstrut\\
    \multicolumn{2}{c}{C=50} \\
    $K=3$ & $N_c=90$ \\
    $K=3$ & $N_c=90$ \Bstrut \\
    \bottomrule
\end{tabular}
\subcaption{CIFAR-10}
\end{subtable}
\caption{Hyperparameters selected for baselines and two variations of ASER. }
\label{tab:hyperparameters}
\end{table*}
}


\section{Discrepancy of CIFAR-10 Result for MIR between Original Paper and our Work}\label{appendix: mir_discrepancy}
{
\setlength\tabcolsep{4.15pt}
\begin{table*}
\begin{subtable}{0.378\textwidth}
    \centering
    \small
    \begin{tabular}{c c c c|} 
    \toprule
    Method &M=1k& M=2k &  M=5k\Bstrut \\
    \hline\hline
    \Tstrut
    MIR&$8.1\pm0.3$&$11.2\pm0.7$&$15.9\pm1.6$\\
    \hline
    MIR$^t$& $7.7\pm0.6$&$10.2\pm0.7$&$17.8\pm1.0$\\
    \hline
    \midrule
    ASER&$11.7\pm0.7$&$\mathbf{14.4\pm0.4}$&$18.2\pm0.7$\\
    ASER$_\mu$&$\mathbf{12.2\pm0.8}$&$14.8\pm1.1$&$18.2\pm1.1$\\
    \hline
    ASER$^t$& $11.3\pm0.4$&$13.8\pm0.5$&$18.0\pm1.0$\\
    ASER$_\mu^t$& $12.2\pm0.5$&$\mathbf{15.3\pm0.5}$&$\mathbf{18.4\pm0.9}$\\
    \hline
    \bottomrule
    \end{tabular}
    \subcaption{Mini-ImageNet}
\end{subtable}
\begin{subtable}{.29\textwidth}
    \centering
\small
\begin{tabular}{c c c |} 
    \toprule
     M=1k&  M=2k &  M=5k \Bstrut \\   
    \hline\hline
    \Tstrut
    $11.2\pm0.3$&$14.1\pm0.2$&$21.2\pm0.6$\\
    \hline
    $11.2\pm0.3$&$14.5\pm0.3$&$21.9\pm0.5$\\
    \midrule
    $12.3\pm0.4$&$14.7\pm0.7$&$20.0\pm0.6$\\
    $\mathbf{14.0\pm0.4}$&$17.2\pm0.5$&$21.7\pm0.5$\\
    \hline
    $13.2\pm0.5$&$16.1\pm0.3$&$20.7\pm0.5$\\
    $13.8\pm0.3$&$\mathbf{17.3\pm0.5}$&$21.5\pm0.7$\\
    \hline
    \bottomrule
\end{tabular}
\subcaption{CIFAR-100}
\end{subtable}
\begin{subtable}{.29\textwidth}
    \centering
\small
\begin{tabular}{c c c} 
    \toprule
    M=0.2k &  M=0.5k & M=1k \Bstrut \\ 
    \hline\hline
    \Tstrut
    $28.3\pm1.6$&$35.6\pm1.2$&$42.4\pm1.5$\\
    \hline
    $28.0\pm1.1$&$36.9\pm1.7$&$44.9\pm0.9$\\
    \midrule
    $27.8\pm1.0$&$36.2\pm1.1$&$43.1\pm1.2$\\
    $26.4\pm1.5$&$36.3\pm1.2$&$43.5\pm1.4$\\
    \hline
    $27.6\pm1.4$&$35.8\pm1.9$&$42.4\pm1.4$\\
    $26.2\pm1.3$&$36.9\pm1.3$&$44.0\pm1.1$\\
    \hline
    \bottomrule
\end{tabular}
\subcaption{CIFAR-10}
\end{subtable}
\caption{The superscript $^t$ means using the "exclude current task samples" trick, namely excluding the samples from current task during \retrieval{}}
\label{tab: trick_result}
\end{table*}
}

In the official repository of MIR, the authors apply a trick to improve performance that is not mentioned in the original paper. Specifically, during \retrieval{}, the trick excludes the memory samples from the current task. Note that to apply this trick, task identity is required during training. Our experimental results for MIR shown in Table \ref{tab:acc_main} are based on the implementation of the original paper and therefore we have not applied this trick.\\ 
\\
To understand the effect of this trick, we apply it to both MIR and our proposed ASER \& ASER$_\mu$. As we can see in Table~\ref{tab: trick_result}, the trick indeed improves the results of MIR, especially in Mini-ImageNet (when M=5k) and CIFAR-10 (when M=1k). Nevertheless, this trick is not always useful. For example, when the memory buffer is small, this trick does not help and, in some cases, shows detrimental effects. In contrast, the trick does not have too much effect on our proposed ASER \& ASER$_\mu$. Most results are very similar to the ones without the trick. 


\end{document}